\documentclass[10pt,twocolumn,letterpaper]{article}

\usepackage{iccv}
\usepackage{times}
\usepackage{epsfig}
\usepackage{graphicx}
\usepackage{amsmath}
\usepackage{amssymb}
\usepackage{bm}
\usepackage{algorithm}
\usepackage{algorithmic}
\usepackage[caption=false,font=footnotesize]{subfig}

\usepackage[breaklinks=true,bookmarks=false]{hyperref}

\iccvfinalcopy 

\def\iccvPaperID{****} 

\begin{document}

\title{A Novel Method for the Absolute Pose Problem with Pairwise Constraints}



\author{
Yinlong Liu\textsuperscript{a}\footnotemark[1],~
Xuechen Li\textsuperscript{b,c}\footnotemark[1],~
Manning Wang\textsuperscript{b,c},~
Guang Chen\textsuperscript{a,d},\\
Zhijian Song\textsuperscript{b,c}\footnotemark[2],~
Alois Knoll\textsuperscript{a}\footnotemark[2]\\
\\
\textsuperscript{a}\emph{Technische Universit\"at M\"unchen}\\
\textsuperscript{b}\emph{Digital Medical Research Center, School of Basic Medical Sciences, Fudan University}\\
\textsuperscript{c}\emph{Shanghai Key Laboratory of Medical Imaging Computing and Computer Assisted Intervention}\\
\textsuperscript{d}\emph{Tongji University}
}
\maketitle
 
\renewcommand{\thefootnote}{\fnsymbol{footnote}} 
\footnotetext[1]{These authors contributed equally to this work.} 
\footnotetext[2]{Corresponding authors: zjsong@fudan.edu.cn,~knoll@in.tum.de}


\maketitle

\begin{abstract}
   Absolute pose estimation is a fundamental problem in computer vision, and it is a typical parameter estimation problem, meaning that efforts to solve it will always suffer from outlier-contaminated data. Conventionally, for a fixed dimensionality $d$ and the number of measurements $N$, a robust estimation problem cannot be solved faster than $\mathcal{O}(N^d )$. Furthermore, it is almost impossible to remove $d$ from the exponent of the runtime of a globally optimal algorithm. However, absolute pose estimation is a geometric parameter estimation problem, and thus has special constraints. In this paper, we consider pairwise constraints and propose a globally optimal algorithm for solving the absolute pose estimation problem. The proposed algorithm has a linear complexity in the number of correspondences at a given outlier ratio. Concretely, we first decouple the rotation and the translation subproblems by utilizing the pairwise constraints, and then we solve the rotation subproblem using the branch-and-bound algorithm. Lastly, we estimate the translation based on the known rotation by using another branch-and-bound algorithm. The advantages of our method are demonstrated via thorough testing on both synthetic and real-world data.  
\end{abstract}

\section{Introduction}

\subsection{Background}
Camera pose estimation is a critical and fundamental problem in computer vision \cite{Campbell2018Globally} and robotics \cite{Grigorescu:2011:RCP:2304781.2304953}. The problem of estimating the absolute pose of a calibrated camera given a certain number of correspondences between 3D world points and 2D image projection points is known as the Perspective-n-Point (PnP) problem \cite{Lepetit2009EP}. It arises as a subtask in many different applications (e.g., robot vision navigation \cite{Taira_2018_CVPR} and camera localization \cite{7572201}).

Mathematically, the absolute pose estimation problem, i.e., the problem of estimating the pose parameters (rotation and translation) given certain observations (3D points and 2D points), is a typical parameter estimation problem \cite{Enqvist:2008:ROP:1478392.1478407}(also a fitting problem \cite{Enqvist:2012:RFM:2402940.2402996}). This problem has been studied for more than a century, and researchers have proposed many methods \cite{Wilcox1997Introduction,8100019,7873535,Cai_2018_ECCV} of improving the solution speed, accuracy, and robustness to outliers. However, a recent study \cite{Chin_2018_ECCV} has shown that fitting a model to data with outliers is an NP-hard problem. A somewhat promising result is that for a fixed dimensionality $d$, a robust estimation problem can be solved in polynomial time in the number of measurements $N$ \cite{Enqvist:2015:TAR:2742226.2742277}. However, this does not imply that a generalized robust estimation problem with outliers can be solved efficiently because a generalized robust fitting problem is a W[1]-hard problem in $d$ dimensions \cite{Chin_2018_ECCV} and, more specifically, it cannot be solved faster than $\mathcal{O}(N^d )$ \cite{7873535,Erickson2006}. Furthermore, it is almost impossible to remove $d$ from the exponent of the run time of a globally optimal algorithm \cite{Chin_2018_ECCV}.

In addition, we can show the "hardness" of the absolute pose estimation with corrupted data from the optimization perspective. Generally, a robust absolute pose estimation problem is always a nonconvex optimization problem \cite{Enqvist:2008:ROP:1478392.1478407,8100019}. There are two reasons why robust absolute pose estimation must be formulated as a “hard” problem. One reason is that the objective function for robust estimation should be robust loss functions, which are always nonconvex functions. The other reason is that the robust estimation problem is optimized in $SE(3)$, which corresponds to two totally different manifolds, rotation ($\textbf{R}\in{SO(3))}$) and translation ($\textbf{t}\in{\mathbb{R}^3)}$). Although there are already some solid theories regarding convex optimization in $\mathbb{R}^3$ \cite{8470970} and $SO(3)$ \cite{convex} separately, robust estimation in $SE(3)$ still seems to be a difficult problem \cite{8100019}. In addition, the dimensionality of $SE(3)$ is six, which increases the hardness of the robust estimation problem. 

In other words, when there are mismatches in the 2D-3D correspondences, the absolute pose estimation problem is a rather “hard” problem. However, in practical applications, outliers are inevitable and will lead to a significant decrease in accuracy for pose estimation \cite{Hartley2013}. Fortunately, the absolute pose estimation problem is a geometric fitting problem and thus may be efficiently solved by considering geometric constraints. 

In this paper, we decouple the rotation and translation subproblems using pairwise constraints, thus reducing the dimensionality of the original problem. Consequently, the original 6-degree-of-freedom (6-DoF) pose estimation problem is transformed into two 3-DoF subproblems, thereby significantly reducing the hardness of the pose estimation problem. As a result, we can efficiently obtain a global solution to the robust pose estimation problem.

\subsection{Related work}

The camera pose problem has been studied for more than a century, and there is a large body of literature on the absolute pose estimation problem \cite{7368948}. Here, we first review the PnP algorithm without mismatches. When the observations include no outliers, the PnP problem has a closed solution ($n\geq3$). To reduce the sensitivity to noise and consider a larger point set,  the Efficient PnP (EPnP) \cite{Lepetit2009EP}, Optimal PnP (OPnP) \cite{6751402}, and Unified PnP (UPnP) \cite{10.1007/978-3-319-10590-1_9} methods have been developed to produce accurate results with a linear complexity. These algorithms are applied in many related areas and can be regarded as state-of-the-art outlier-free PnP techniques.

When the observations includes outliers, the most commonly applied mechanism is RANdom SAmple Consensus (RANSAC) \cite{Fischler:1981:RSC:358669.358692,6365642}, which is a well-known algorithm for robust parameters estimation that is widely used for the camera pose estimation problem. However, as its name suggests, it is a nondeterministic heuristic algorithm, which means that RANSAC provides no guarantee regarding the optimality of its solution. The most recent advancement is to remove outliers before applying an outlier-free PnP method. In the Robust Efficient Procrustes PnP method(REPPnP) \cite{6909465}, the pose estimation problem is formulated as a low-rank homogeneous system, and outliers are iteratively removed under the assumption that the rank of the null space of the linear system should always be one.  Re-weighting and 1-Point RANSAC-based PnP (R1PPnP) \cite{8470970} uses a heuristic method of handling outliers by utilizing a soft reweighting mechanism and the 1-point RANSAC scheme. However, the outlier removal problem is as hard as the original problem. Nevertheless, although it is difficult to eliminate all outliers efficiently, the proportion of outliers in the observations can be reduced using these outlier removal methods. Moreover, in practical applications, it may be possible to obtain prior knowledge that can be used in outlier removal. For example, \cite{Camposeco_2017_CVPR} presents an outlier filter that incorporates prior information on the viewing directions, \cite{7534854} presents an approximate outlier rejection scheme with a known vertical direction, and the method proposed in \cite{Larsson2016Outlier} requires knowledge about the overall camera orientation with which to prune outliers.

In addition to the PnP methods discussed above, there is another class of methods for solving the robust pose estimation problem. In this body of work, the pose estimation problem is formulated as a robust optimization problem. M-estimator \cite{7368948} is a classical robust estimation method, but it always solves to a local optimum because of the nonconvexity of the objective function. Therefore, more recent work on robust estimation has focused on obtaining globally optimal solutions. The most popular algorithm may be the branch-and-bound algorithm, which is always combined with convex relaxation \cite{8100019,4531744,Briales_2017_CVPR} or geometric relaxation \cite{Enqvist:2008:ROP:1478392.1478407,Hartley:2009:GOT:1502536.1502557}. However, the branch-and-bound-based algorithms devoted to pose estimation always suffer from a heavy computational burden for the obvious reason that the dimensionality of the feasible domain for pose estimation is six, and thus, the pose estimation problem perhaps cannot be regarded as a low-dimensional optimization problem from the perspective of using branch-and-bound approach. In other words, even if the branch-and-bound algorithm has tight bounds, it still needs considerable time to search the entire feasible space in $SE(3)$. Moreover, the optimization is performed in two totally different manifolds, and it is not easy to calculate a tight bound for each branch.

Another topic that is closely related to the absolute pose estimation problem is point set registration \cite{Campbell2018Globally,yang2016goicp}. The only difference is that in the point set registration, there is no existing point correspondence. Similarly, the search for globally optimal solutions is a hot topic in the field of point set registration, and the branch-and-bound algorithm has also been broadly applied in recent related studies. One of the most successful algorithms for this purpose may be the algorithm proposed in \cite{4409077} and its subsequent versions \cite{Campbell2018Globally,yang2016goicp,Bustos:2014:FRS:2679600.2679783}. These works are all based on rotation search theory, and for $SE(3)$ optimization in particular, a more systematic scheme called the nested branch-and-bound is applied. Moreover, the decoupling methods presented in \cite{8099741} improve efficiency, which inspires us to decouple the rotation and translation subproblems by means of pairwise constraints. 

\subsection{Contributions of this paper}
In this paper, we introduce a novel robust and global solution to the absolute pose estimation problem, called Robust and Global PnP (RGPnP). The contributions are three-fold. (1) Our proposed method produces a globally optimal solution to the absolute pose problem. We apply the branch-and-bound algorithm, which is a global optimization algorithm, to obtain the best solution. (2) We use novel pairwise constraints to decouple the rotation and translation subproblems, which can then be efficiently solved sequentially. (3) The proposed method is robust to outliers. We use a robust objective function, namely, consensus maximization, which can be regarded as a 0-1 loss function and has already been successfully used in many robust fitting problems. 

\section{Method}

\subsection{Problem formulation}
In this paper, we formulate the absolute pose estimation problem as follows. Let the $i$-th 3D points in the world coordinate system be denoted by $\emph{\textbf{p}}_i\in{\mathbb{R}^3}$, $i=1,\ldots, n$. Similarly, let $\emph{\textbf{q}}_i\in{\mathbb{S}^2}$, $i=1,\ldots, n$ be the $i$-th bearing vector with a unit norm, which corresponds to the $i$-th 2D point in the camera coordinate system. $\textbf{R}\in{SO(3)}$ is the rotation and $\textbf{t}\in{\mathbb{R}^3}$ is the translation. Given these definitions, the relationship for inlier observation is as follows:
\begin{equation}
\lambda_i\emph{\textbf{q}}_i=\textbf{R}\emph{\textbf{p}}_i+\textbf{t}, i=1, \ldots, n
\end{equation}
where $\lambda_i$ is the unknown depth of the $i$-th point. The objective of the absolute pose estimation problem is then to estimate the rotation and translation, given $n$ pairs of points. Alternatively, to eliminate $\lambda_i$, eq(1) can be reformulated as follows:
\begin{equation}
\angle(\emph{\textbf{q}}_i, \textbf{R}\emph{\textbf{p}}_i+\textbf{t})=0
\end{equation}
where $\angle(\emph{\textbf{a}}, \emph{\textbf{b}})$ is the angle between vectors \emph{\textbf{a}} and \emph{\textbf{b}}. In this paper, we estimate the camera pose by maximizing the cardinality $E$ of the inlier set $\mathcal{S}_I$:
\begin{equation}
E^*(\textbf{R},\textbf{t})=max|\mathcal{S}_I\vert
\end{equation} 
\begin{equation}
\mathcal{S}_I=\left\{(\emph{\textbf{q}}_i, \emph{\textbf{p}}_i)|\angle(\emph{\textbf{q}}_i, \textbf{R}\emph{\textbf{p}}_i
+\textbf{t})<\epsilon\right\}
\end{equation}
where $\epsilon$ is the inlier threshold.

The function given in eq(4) is inherently robust to outliers since matched points are considered inliers only if their angular separation is below the inlier threshold $\epsilon$. However, obtaining the global solution to eq(3) is a nontrivial problem. We propose the use of a set of novel pairwise constraints to obtain an equivalent but easier problem.  

\subsection{Eliminating translation by means of pairwise constraints}
We consider two pairs of inlier correspondences $(\emph{\textbf{p}}_i,\emph{\textbf{q}}_i)$ and $(\emph{\textbf{p}}_j,\emph{\textbf{q}}_j)$. When they are aligned as shown in Fig.1, the four points and the center of the camera must all lie in the same plane. $\emph{\textbf{v}}=\emph{\textbf{q}}_i\times\emph{\textbf{q}}_j$  is the normal of that plane, and $\emph{\textbf{l}}=(\textbf{R}\emph{\textbf{p}}_i+\textbf{t})-(\textbf{R}\emph{\textbf{p}}_j+\textbf{t})=\textbf{R}(\emph{\textbf{p}}_i-\emph{\textbf{p}}_j)$ is a vector in the plane. For simplicity, let $\emph{\textbf{u}}=\emph{\textbf{p}}_i-\emph{\textbf{p}}_j$; then, $\emph{\textbf{l}}=\textbf{R}\emph{\textbf{u}}$. The relation between $\emph{\textbf{u}}$ and $\emph{\textbf{v}}$ is obvious: $\emph{\textbf{v}}\perp\textbf{R}\emph{\textbf{u}}$ or $\emph{\textbf{v}}^{\textbf{T}}\bm{\cdot}\textbf{R}\emph{\textbf{u}}=0$, which is called a pairwise constraint in this paper (note that a similar equation was used in \cite{Ke_2017_CVPR}). Such pairwise constraints provide an easy and elegant yet powerful means of decoupling the rotation and translation in eq(3); thus we can calculate the optimal rotation first by enforcing these pairwise constraints. Consequently, we define a new objective function with only rotation parameters as follows:
\begin{equation}
\mathcal{Q}^*(\textbf{R})=max|\mathcal{S}^p_I\vert
\end{equation}
\begin{equation}
\mathcal{S}^p_I=\left\{\left(\emph{\textbf{q}}_i, \emph{\textbf{q}}_j, \emph{\textbf{p}}_i, \emph{\textbf{p}}_j\right)|
|\angle\left(\emph{\textbf{q}}_i\times\emph{\textbf{q}}_j, \textbf{R}(\emph{\textbf{p}}_i-\emph{\textbf{p}}_j)\right)-\frac{\pi}{2}\vert<\delta\right\}
\end{equation}
where $i\neq j$ and $\delta$ is a new inlier threshold.

We can also use $(\emph{\textbf{u}},\emph{\textbf{v}})$ to rewrite eq(5) as follows:
\begin{equation}
\mathcal{Q}^*(\textbf{R})=max\sum_k\lfloor |\angle(\emph{\textbf{v}}_k, \textbf{R}\emph{\textbf{u}}_k)-\frac{\pi}{2}\vert<\delta\rfloor
\end{equation}
Here, $\lfloor \bm{\cdot} \rfloor$ is a 0-1 function that returns a value of 1 if the condition is true and a value of 0 otherwise, and $k=1, \ldots, m$ is the index of the $(\emph{\textbf{u}},\emph{\textbf{v}})$ pairs. 

\begin{figure}[tbp]
\centering
\includegraphics[width=3.5in]{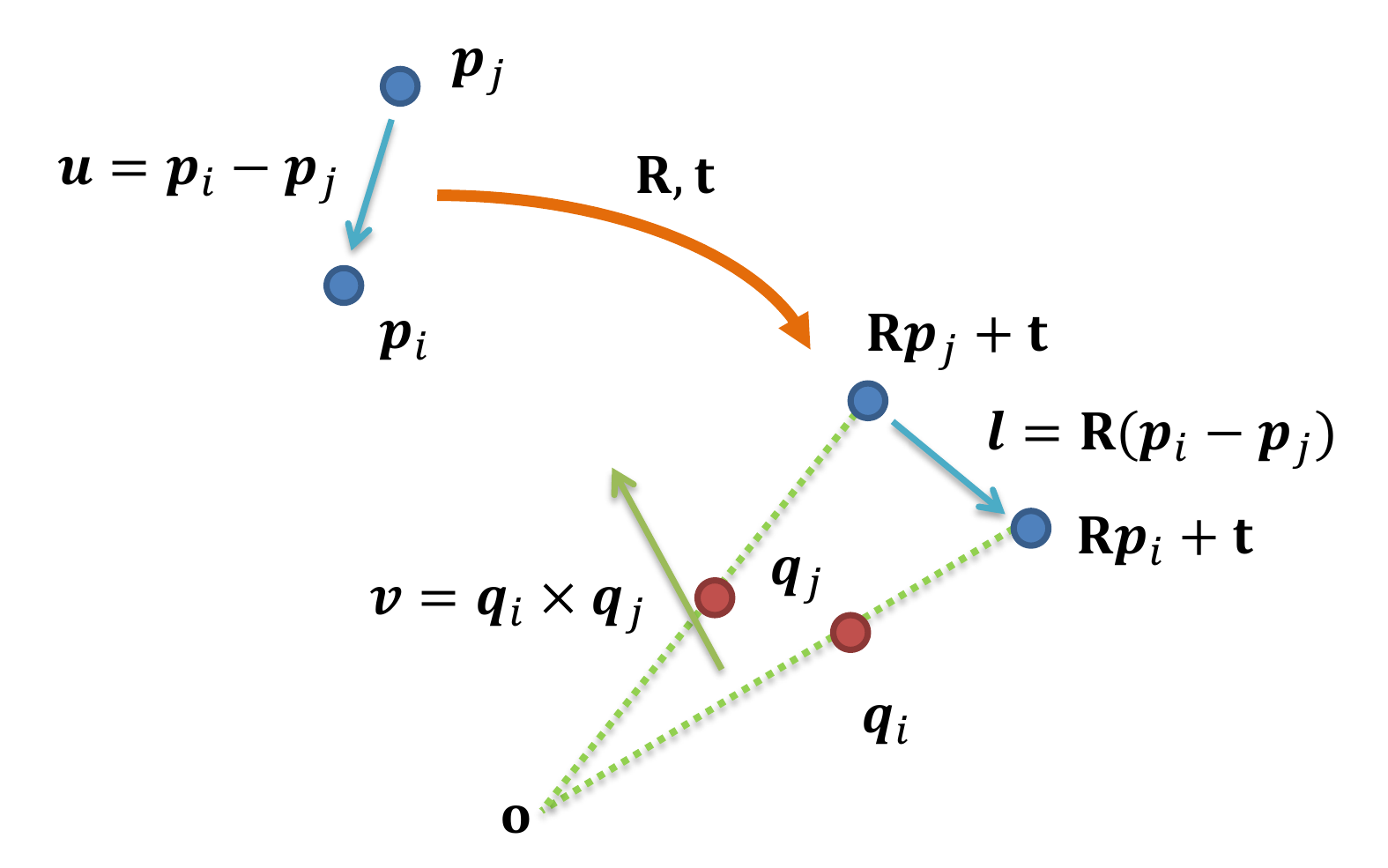}
   \caption{Geometric relations and pairwise constraints between a pair of 3D points (blue) and their corresponding 2D points (red).}
\end{figure}

We find that there is only the rotation to be solved for in eq(7); thus, we have already successfully reduced the 6-DoF pose problem to a 3-DoF rotation estimation problem in $SO(3)$. However, the number of input data increases from $n$ to $0.5n(n-1)$. In \cite{7755788}, the authors pointed out that estimated parameters can be found as a solution on a subset of all the input data. Unfortunately, the number of $(\emph{\textbf{u}},\emph{\textbf{v}})$ pairs is very large, and there are many different ways of choosing a subset from all samples. For a parameter estimation problem, all original input observations are expected to be involved in the  estimation. Interestingly, we find that if each original correspondence is used once, then the number of $(\emph{\textbf{u}},\emph{\textbf{v}})$ pairs decreases to $0.5n$ under our pairwise constraints. If the outlier ratio is not very large, we recommend using this $0.5n$-subset as the input so that every original observation is involved in the estimation. However, if the outlier ratio is large, we recommend increasing the input size. 

\subsection{Global $\textbf{SO(3)}$ search}
In this section, we introduce a method based on the branch-and-bound algorithm for obtaining the global solution to eq(7). We summarize the proposed method in \textbf{Algorithm 1}. In brief, the branch-and-bound algorithm proceeds by recursively subdividing and pruning the rotation space until the global optimum is found. In this paper, the rotation space $SO(3)$ is minimally parameterized with an angle-axis representation, and a 3D cube with a side length of $2\pi$ is used as the rotation domain. For more details about the angle-axis representation, please refer to \cite{Hartley:2009:GOT:1502536.1502557}.

Generally, the success of a branch-and-bound algorithm depends on the quality of its upper and lower bounds. In this paper, we present two different ways to calculate the bounds. The first pair of bounds is derived based on Hartley and Kahl$'$s rotation search theory. To obtain the second pair of bounds, the rotation matrix is stacked into a $9\times1$ vector and eq(7) is reformulated as a linear system.

\begin{algorithm}
\caption{Branch-and-bound algorithm for obtaining the rotation}
\label{alg1}
\begin{algorithmic}[1]
\REQUIRE Correspondence pairs $\left\{(\emph{\textbf{v}}_k, \emph{\textbf{u}}_k )\right\}_{k=1}^m$ and inlier threshold. 
\STATE Initialize $\mathbb{B}\leftarrow$ cube of side length $2\pi$, and insert $\mathbb{B}$ into a priority queue $q$.
\WHILE {$q$ is not empty}
\STATE Subdivide $\mathbb{B}$ into eight cubes $\left\{\mathbb{B}_d\right\}_{d=1}^8$.
\STATE For each $\mathbb{B}_d$ calculate the upper and lower bounds $\left\{\mathcal{Q}^u_d, \mathcal{Q}^l_d\right\}_{d=1}^8$.
\STATE Update the best solution so far: $\mathcal{Q}^*(\textbf{R}^*)=max\left\{\mathcal{Q}_i^l\right\}$, $i$ for all branches.
\STATE Remove the branches that $\mathcal{Q}^u_i<\mathcal{Q}^*$, $i$ for all branches.
\STATE Update the highest priority cube $\mathbb{B}$ with upper bound $\mathcal{Q}^u$ for the next loop.
\IF {$\mathcal{Q}^u=\mathcal{Q}^*$}
   \STATE terminate and return $\textbf{R}^*$.
\ENDIF 
\ENDWHILE
\RETURN Optimal rotation $\textbf{R}^*$.
\end{algorithmic}
\end{algorithm}

\subsubsection*{Bounds from Hartley and Kahl's theoty}
Let us start with a famous equation that was proved in \cite{Hartley:2009:GOT:1502536.1502557}. Given a cube-shaped branch $\mathbb{B}$ of the rotation space, whose center is $\textbf{R}_0$, for any $\emph{\textbf{u}}\in\mathbb{R}^3$ and any $\textbf{R}\in\mathbb{B}$, the following holds:
\begin{equation}
\angle(\textbf{R}\emph{\textbf{u}}, \textbf{R}_0\emph{\textbf{u}})\leq\sqrt{3}\sigma
\end{equation}
where $\sigma$ is the half-side length of the cube $\mathbb{B}$. According to the triangle inequality in a spherical geometry, for any $\emph{\textbf{v}}\in{\mathbb{R}^3}$
\begin{align}
\angle(\emph{\textbf{v}},\textbf{R}\emph{\textbf{u}})
&\leq\angle(\emph{\textbf{v}},\textbf{R}_0\emph{\textbf{u}})+\angle(\textbf{R}\emph{\textbf{u}},\textbf{R}_0\emph{\textbf{u}})\\
&\leq\angle(\emph{\textbf{v}},\textbf{R}_0\emph{\textbf{u}})+\sqrt{3}\sigma
\end{align}
\begin{align}
\angle(\emph{\textbf{v}},\textbf{R}\emph{\textbf{u}})
&\geq\angle(\emph{\textbf{v}},\textbf{R}_0\emph{\textbf{u}})-\angle(\textbf{R}\emph{\textbf{u}},\textbf{R}_0\emph{\textbf{u}})\\
&\geq\angle(\emph{\textbf{v}},\textbf{R}_0\emph{\textbf{u}})-\sqrt{3}\sigma
\end{align}

From eq(9)-eq(12), for a given pair $(\emph{\textbf{u}}_k, \emph{\textbf{v}}_k)$, we can obtain
\begin{equation}
|\angle(\emph{\textbf{v}}_k, \textbf{R}\emph{\textbf{u}}_k)-\frac{\pi}{2}\vert\geq
|\angle(\emph{\textbf{v}}_k, \textbf{R}_0\emph{\textbf{u}}_k)-\frac{\pi}{2}\vert-\sqrt{3}\sigma
\end{equation}
Then,
\begin{equation}
\lfloor|\angle(\emph{\textbf{v}}_k, \textbf{R}\emph{\textbf{u}}_k)-\frac{\pi}{2} \vert<\delta\rfloor
\leq\lfloor|\angle(\emph{\textbf{v}}_k, \textbf{R}_0\emph{\textbf{u}}_k)-\frac{\pi}{2}\vert-\sqrt{3}\sigma<\delta\rfloor
\end{equation}
As a result, the upper bound of $\mathcal{Q}^*(\textbf{R})$ in eq(7) for any $\textbf{R}\in{\mathbb{B}}$ is 
\begin{equation}
\mathcal{Q}^{upper}_H(\mathbb{B})=
\sum_k\lfloor|\angle(\emph{\textbf{v}}_k, \textbf{R}_0\emph{\textbf{u}}_k)-\frac{\pi}{2}\vert<\delta+\sqrt{3}\sigma\rfloor
\end{equation}
The lower bound can be easily calculated as follows:
\begin{equation}
\mathcal{Q}^{lower}_H(\mathbb{B})=\sum_k\lfloor|\angle(\emph{\textbf{v}}_k, \textbf{R}_0\emph{\textbf{u}}_k)-\frac{\pi}{2}\vert<\delta\rfloor
\end{equation}
The proof for lower bound is obvious because no rotation in the branch can be no better than the optimum.

\subsubsection*{Bounds derived from a linear system formulation}
From the equation $\emph{\textbf{v}}^T\bm{\cdot}\textbf{R}\emph{\textbf{u}}=0$, we can obtain the linear homogeneous equation $\emph{\textbf{e}}^T\emph{\textbf{x}}=0$, where $\emph{\textbf{x}}^T=(\textbf{R}_{1,1}, \textbf{R}_{2,1},\ldots, \textbf{R}_{3,3})$ and $\emph{\textbf{e}}^T=(v_{1}u_{1}, v_{2}u_{1},\ldots, v_{3}u_{3})$. Then, we have another orthogonal relation, $\angle(\emph{\textbf{e}}, \emph{\textbf{x}})=\frac{\pi}{2}$, and we can reformulate eq(7) as
\begin{equation}
\mathcal{Q}^*(\textbf{R})=max\sum_k\lfloor|\angle(\emph{\textbf{e}}_k, \emph{\textbf{x}})-\frac{\pi}{2}\vert<\tau\rfloor
\end{equation}
where $\tau$ is a different new inlier threshold. Notably, eq(17) is the outlier-robust form of the linear system $E\emph{\textbf{x}}=0$, where $E^T=(\emph{\textbf{e}}_1, \emph{\textbf{e}}_2,\ldots, \emph{\textbf{e}}_m)$.

To derive the upper bound of eq(17), we introduce the famous lemma 2 in \cite{Hartley:2009:GOT:1502536.1502557}, which states that the angular distance between two rotations is less than the Euclidean distance between them in the angle-axis representation:
\begin{equation}
\angle(\textbf{R}_1, \textbf{R}_2)\leq\|\emph{\textbf{r}}_1-\emph{\textbf{r}}_2\Vert
\end{equation}
where $\textbf{R}_1$ and $\textbf{R}_2$ are two rotations and $\emph{\textbf{r}}_1$ and $\emph{\textbf{r}}_2$, respectively, are their angle-axis representations. Additionally, according to \cite{Huynh:2009:MRC:1574521.1574531},
\begin{equation}
trace(\textbf{R}^T_1\textbf{R}_2)=1+2cos(\angle(\textbf{R}_1, \textbf{R}_2))
\end{equation}
Meanwhile,
\begin{equation}
trace(\textbf{R}^T_1\textbf{R}_2)=\emph{\textbf{x}}^T_1\emph{\textbf{x}}_2
\end{equation}
where $\emph{\textbf{x}}_1$ and $\emph{\textbf{x}}_2$ are the linear representations of $\textbf{R}_1$ and $\textbf{R}_2$, respectively. Then
\begin{align}
\angle(\emph{\textbf{x}}_1, \emph{\textbf{x}}_2) 
&=cos^{-1}\left(\frac{1}{\|\emph{\textbf{x}}_1\Vert\|\emph{\textbf{x}}_2\Vert}\emph{\textbf{x}}^T_1\emph{\textbf{x}}_2\right)\\
&=cos^{-1}\left(\frac{1}{3}\emph{\textbf{x}}^T_1\emph{\textbf{x}}_2\right) \\
&=cos^{-1}\left(\frac{1}{3}trace(\textbf{R}^T_1\textbf{R}_2)\right) \\
&=cos^{-1}\left(\frac{1}{3}\left(1+2cos\left(\angle(\textbf{R}_1, \textbf{R}_2)\right)\right)\right) \\
&\leq cos^{-1}\left(\frac{1}{3}\left(1+2cos(\|\emph{\textbf{r}}_1-\emph{\textbf{r}}_2\Vert)\right)\right) 
\end{align}

Eq(25) establishes a relation between the angle-axis representation and the linear representation. Geometrically, a cube-shaped branch in the angle-axis representation can be relaxed to a continuous region in the linear representation, as shown in Fig.2.

\begin{figure}[tbp]
\centering
\subfloat[]{\includegraphics[width=2.8in]{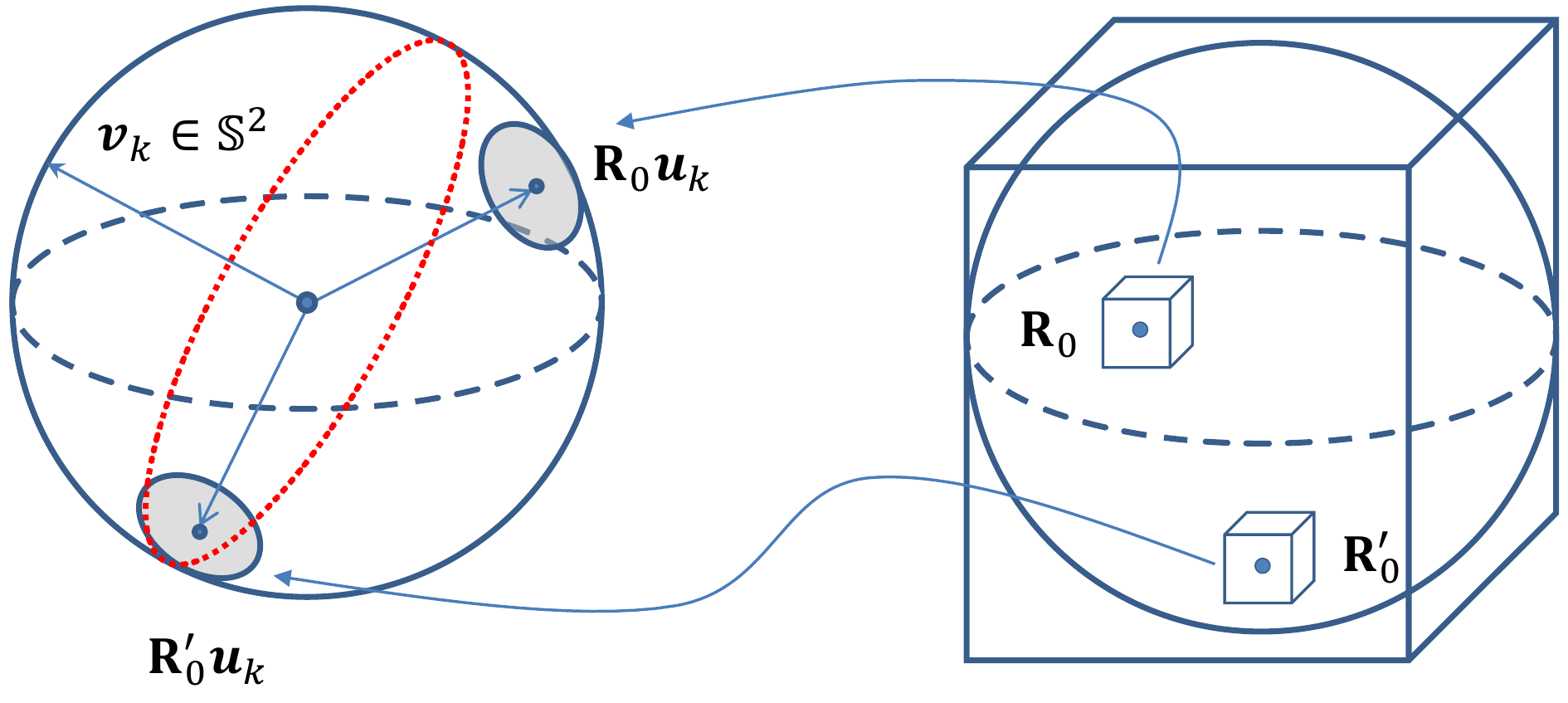}
\label{fig_first_case}}
\hfil
\subfloat[]{\includegraphics[width=3in]{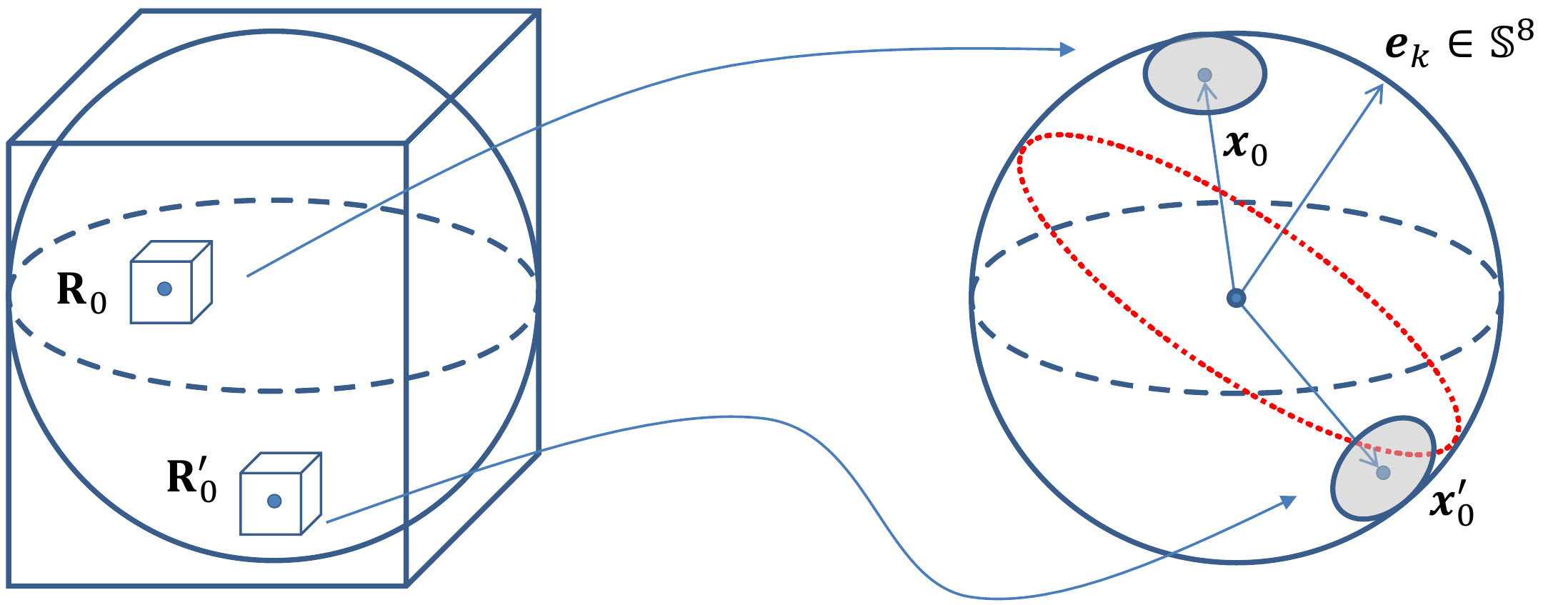}
\label{fig_second_case}}
\caption{Geometric interpretation. (a) The geometric interpretation of the first bound: under the action of all possible rotations within a cube in the angle-axis representation, a unit vector may lie only on a spherical patch on the 3D unit sphere. (b) The geometric interpretation of the second bound: a cube in the angle-axis representation can be mapped to a continuous domain in $\mathbb{S}^8$.}
\end{figure}

Specifically, in a cube-shaped branch $\mathbb{B}$ whose center is $\textbf{R}_0$ (where $\emph{\textbf{x}}_0$ and $\emph{\textbf{r}}_0$ are the linear and angle-axis representations, respectively, of $\textbf{R}_0$), for any $\textbf{R}\in{\mathbb{B}}$,
\begin{align}
\angle(\emph{\textbf{x}}, \emph{\textbf{x}}_0)
&\leq cos^{-1}\left(\frac{1}{3}(1+2cos(\|\emph{\textbf{r}}-\emph{\textbf{r}}_0\Vert))\right)\\
&\leq cos^{-1}\left(\frac{1}{3}\left(1+2cos(\sqrt{3}\sigma)\right)\right)=\alpha
\end{align}
where $\emph{\textbf{x}}$ and $\emph{\textbf{r}}$ are the linear and angle-axis representations, respectively, of $\textbf{R}$; $\sigma$ is the half-side length of the cube $\mathbb{B}$; and $\alpha$ denotes the upper bound of $\angle(\emph{\textbf{x}}, \emph{\textbf{x}}_0)$. Similar to the first bound, we have
\begin{equation}
\angle(\emph{\textbf{e}}_k, \emph{\textbf{x}})\leq\angle(\emph{\textbf{e}}_k, \emph{\textbf{x}}_0)+\angle(\emph{\textbf{x}}, \emph{\textbf{x}}_0)\leq\angle(\emph{\textbf{e}}_k, \emph{\textbf{x}}_0)+\alpha
\end{equation}
\begin{equation}
\angle(\emph{\textbf{e}}_k, \emph{\textbf{x}})\geq\angle(\emph{\textbf{e}}_k, \emph{\textbf{x}}_0)-\angle(\emph{\textbf{x}}, \emph{\textbf{x}}_0)\geq\angle(\emph{\textbf{e}}_k, \emph{\textbf{x}}_0)-\alpha
\end{equation}
Then,
\begin{equation}
|\angle(\emph{\textbf{e}}_k, \emph{\textbf{x}})-\frac{\pi}{2}\vert\geq|\angle(\emph{\textbf{e}}_k, \emph{\textbf{x}})-\frac{\pi}{2}\vert-\alpha
\end{equation}
\begin{equation}
\Rightarrow\lfloor|\angle(\emph{\textbf{e}}_k, \emph{\textbf{x}})-\frac{\pi}{2}\vert<\tau\rfloor\leq\lfloor|\angle(\emph{\textbf{e}}_k, \emph{\textbf{x}})-\frac{\pi}{2}\vert-\alpha<\tau\rfloor
\end{equation}
The upper bound can be derived as
\begin{equation}
\mathcal{Q}^{upper}_L(\mathbb{B})=\sum_k\lfloor|\angle(\emph{\textbf{e}}_k, \emph{\textbf{x}}_0)-\frac{\pi}{2}\vert<\tau+\alpha\rfloor
\end{equation}
The lower bound can be estimated as shown in eq(33), which is similar to eq(16)
\begin{equation}
\mathcal{Q}^{lower}_L(\mathbb{B})=\sum_k\lfloor|\angle(\emph{\textbf{e}}_k, \emph{\textbf{x}}_0)-\frac{\pi}{2}\vert<\tau\rfloor
\end{equation}

Now, we have two types of bounds for objective function within a certain feasible domain. Because they have different formulations, it is very difficult to compare these two pairs of bounds theoretically. However, experiments show that the first formulation based on Hartley and Kahl$'$s theory, is more efficient.

\subsection{Global translation search}
Once the optimal rotation has been obtained, the problem becomes a subproblem of robust absolute pose estimation with a known orientation \cite{Larsson2016Outlier}. In this paper, we introduce an efficient method of solving the translation subproblem via three one-dimensional optimizations rather than one three-dimensional optimization.  

First, we use the known rotation to reduce the outlier ratio. For a pair of correspondences, both correspondences will be considered outliers if they do not satisfy the pairwise constraint shown in eq(34).
\begin{equation}
|\angle(\emph{\textbf{q}}_i\times\emph{\textbf{q}}_j, \textbf{R}(\emph{\textbf{p}}_i-\emph{\textbf{p}}_j))-\frac{\pi}{2}\vert<\delta, i\neq j
\end{equation}

Notably, when the input is the $0.5n$-subset and an inlier and an outlier are paired, the inlier and outlier are both discarded. If the outlier ratio is small, we can still find the solution to the original problem from the remaining data. However, if the outlier ratio is large, we recommend increasing the input, e.g., pairing each correspondence with more than one other correspondence, to preserve as many inliers as possible. The reason is apparent: despite the discarding of inlier-and-outlier pairs, the same inliers are likely to be present in other pairs with other inliers. Theoretically, this step cannot remove all outliers, but it will significantly reduce the number of outliers .

The next step is to calculate the translations from each pair constructed from the remaining correspondences. We will then have many translations, which will include some false results. Next, we must find the best translation among these translation results, for which the best solution can be obtained by voting based on the branch-and-bound algorithm. Moreover, a translation is defined by three independent variables and we can optimize those three variables independently. Consequently, the dimensionality of the problem decreases from three to one. For the one-dimensional branch-and-bound method, we formulate the objective function as shown in eq(35)
\begin{equation}
\mathcal{T}^*=max\sum_s\lfloor|\textbf{t}-\textbf{t}_s\vert\leq\varepsilon\rfloor
\end{equation}
where $\textbf{t}_s$ is the $s$-th solution and $\varepsilon$ is the inlier threshold. The search domain is easily determined: $\textbf{t}\in{\left[min(\textbf{t}_s), max(\textbf{t}_s)\right]}$. Given the divided domain, whose center is $\textbf{t}_0$ and whose half-side length is $\mu$, the upper and lower bounds are as follows:
\begin{equation}
\mathcal{T}_u=\sum_s\lfloor|\textbf{t}_0-\textbf{t}_s\vert\leq\varepsilon+\mu\rfloor
\end{equation}
\begin{equation}
\mathcal{T}_l=\sum_s\lfloor|\textbf{t}_0-\textbf{t}_s\vert\leq\varepsilon\rfloor
\end{equation}

\section{Experiments}
In this section, we report the results of evaluating our method on both synthetic and real-word data. To highlight the contributions of this study, all experiments were conducted with various outlier ratios, while outlier-free cases are not considered here. Based on the two different types of bounds derived in Sec 2.3, the two versions of the methods proposed in this paper are denoted by RGPnP$\_$H (Hartley and Kahl$’$s theory) and RGPnP$\_$L (linear system formulation). Here, the input set of pairs of correspondences is the $0.5n$-subset, as described in Sec 2.2, for all experiments. The proposed methods were compared against several baseline approaches, including RANSAC+P3P with a maximum of 1000 trials (RNSC1000+P3P) and 5000 trials (RNSC5000+P3P), REPPnP \cite{6909465}, and R1PPnP \cite{8470970}, of which the latter two methods can be regarded as state-of-the-art methods of handling the absolute pose estimation problem with outliers. All experiments were conducted using MATLAB 2018b on a computer equipped with a 3.2 GHz Intel Xeon E5  CPU.

\subsection{Experiments with synthetic data}
For synthetic experiments, we assumed a camera with an image size of $640\times480$ and a focal length of 1000 pixels. We randomly generated 1000 3D points in a cubic region of $[0,10]\times[0,10]\times[5,15]$ and projected them onto the image to generate correct correspondences. Outliers were added to both the 3D points and 2D images to generate incorrect matches. Two different types of outliers were added, as follows: (1) Uniformly distributed 3D points were generated in the same cube as the data points ($[0,10]\times[0,10]\times[5,15]$), and each of them was assigned a correspondence to a randomly generated 2D points in the image. (2) Uniformly distributed 3D points were generated in a cubic region of $[0,1]\times[0,1]\times[0,1]$, different from the region of the data points, and each of them was assigned a correspondence to a randomly generated point in the image. The outlier ratio is defined as $r_{outlier}=\frac{N_{outlier}}{N_{outlier}+N_{inlier}}$. We performed experiments with different outlier ratios, and for each ratio, 500 trials were run for each method.

To evaluate the estimation accuracy, we computed the rotation error in degrees between the ground-truth rotation $\textbf{R}_{true}$ and the estimated $\textbf{R}$ as $e_{rot}=\angle(\textbf{R}_{true}, \textbf{R})$ and the translation error between the ground-truth translation $\textbf{t}_{true}$ and the estimated $\textbf{t}$ as $e_{trans}=\frac{\|\textbf{t}_{true}-\textbf{t}\Vert}{\|\textbf{t}_{true}\Vert}\times100\%$. We report the success rate, defined as the fraction of trials in which the correct pose was found, where an estimation was considered successful when $e_{rot}$ was less than 0.1 radius and $e_{trans}$ was less than 0.2. The success rates for 500 trials of each method for both types of outliers are plotted in Fig.3.

\begin{figure}[tbp]\setcounter{subfigure}{-1}
\subfloat{\includegraphics[width=3.2in]{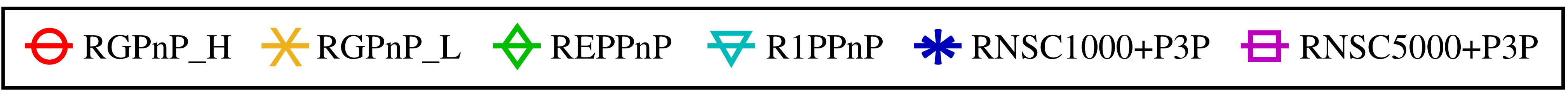}
\label{fig_legend}}\\
\centering
\subfloat[the first type of outliers]{\includegraphics[width=1.6in]{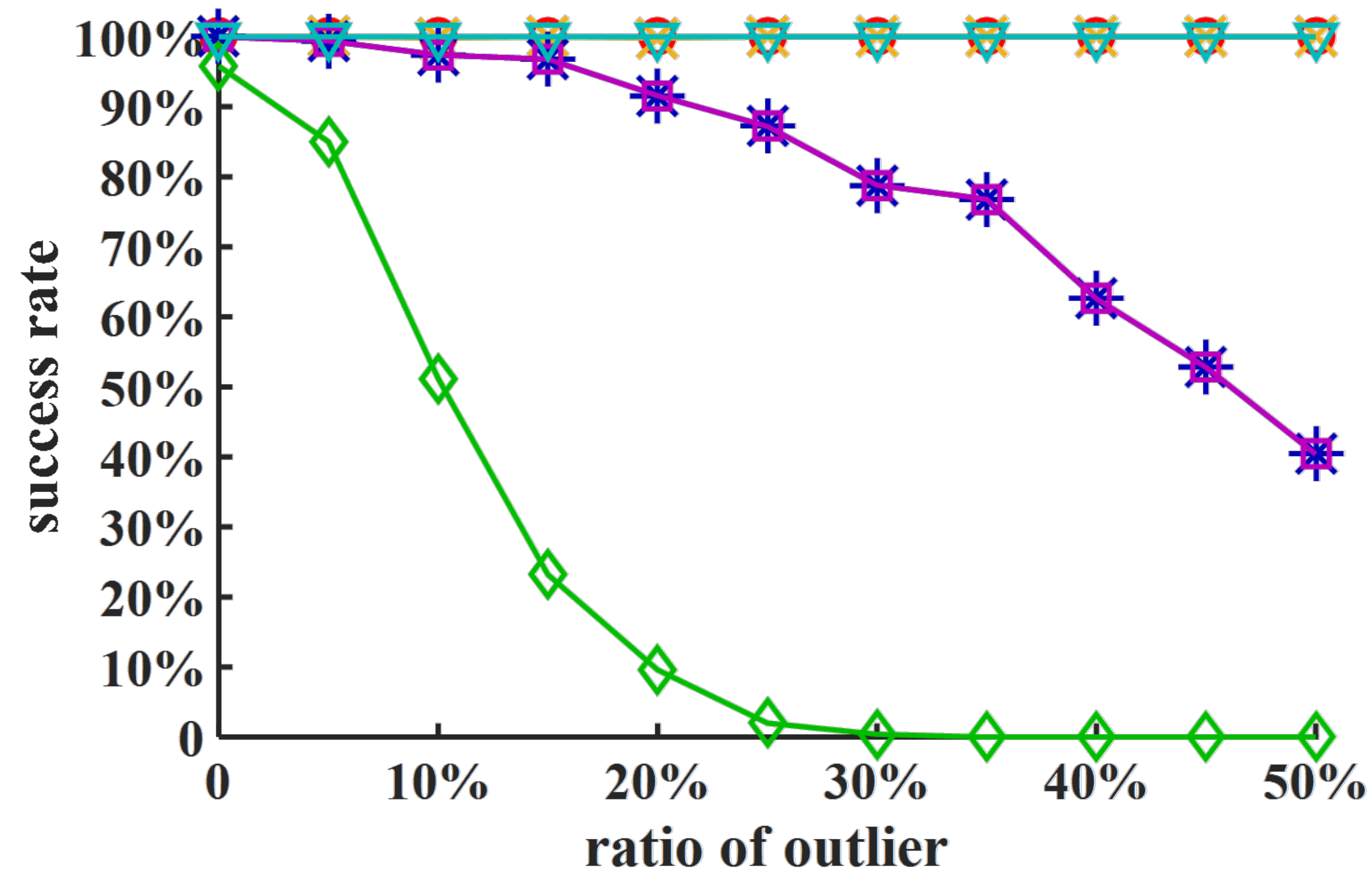}
\label{fig_first_case}}
\hfil
\subfloat[the second type of outliers]{\includegraphics[width=1.6in]{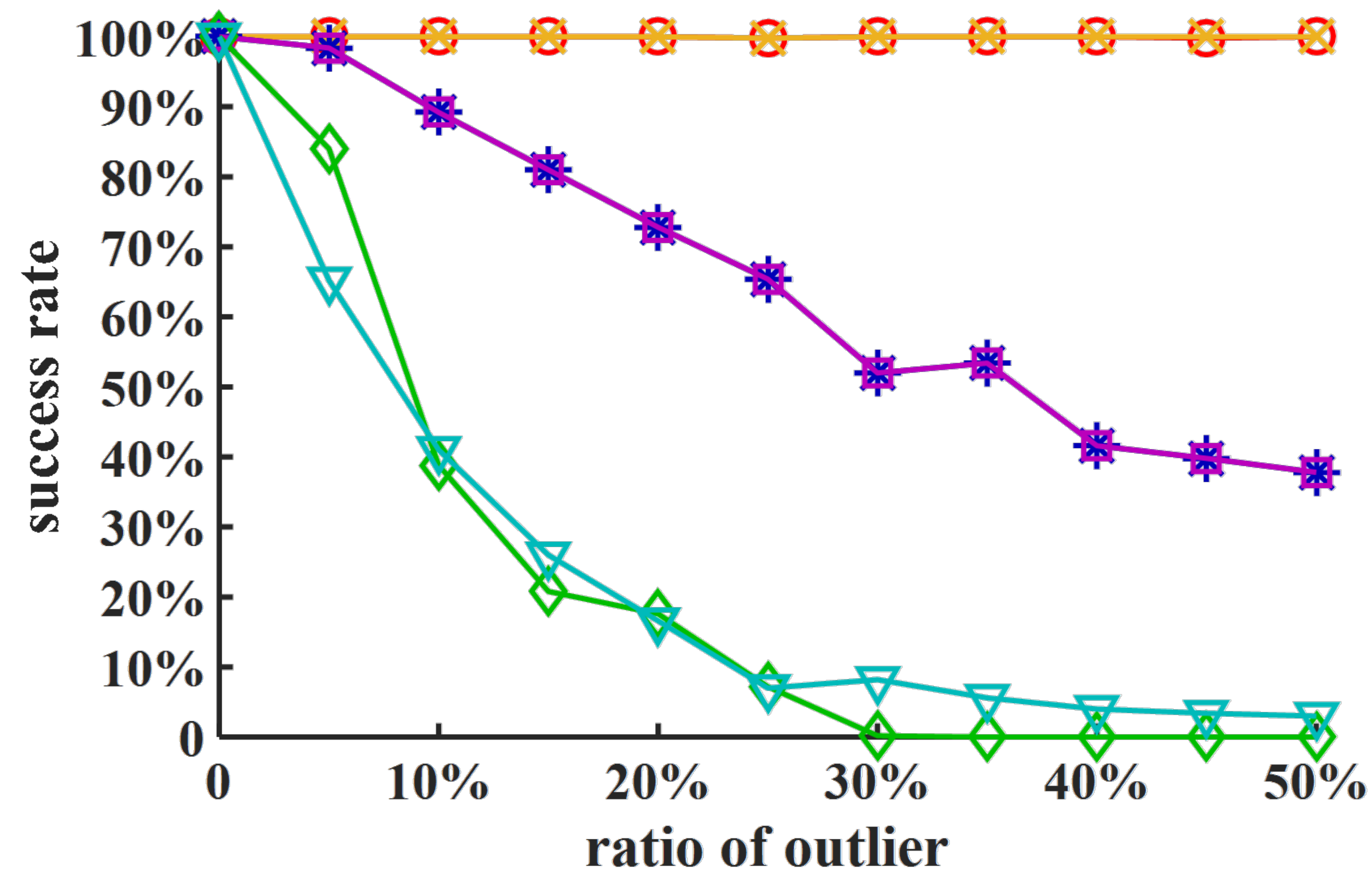}
\label{fig_second_case}}
\caption{ Success rates for both types of outliers.}
\end{figure}

\begin{figure*}[htbp]
\centering
\subfloat[]{\includegraphics[width=2.25in]{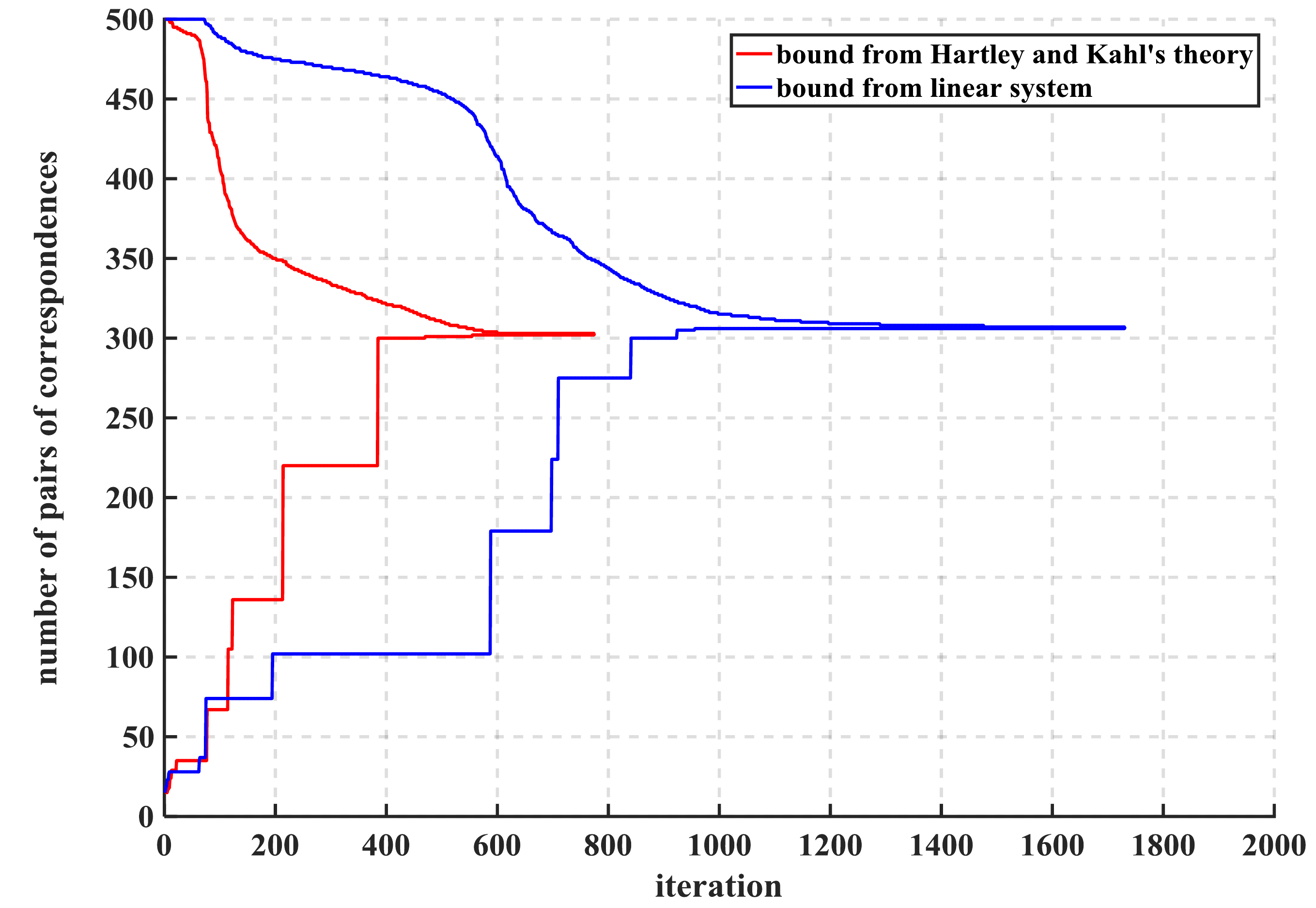}
\label{fig_first_case}}
\hfil
\subfloat[]{\includegraphics[width=2.25in]{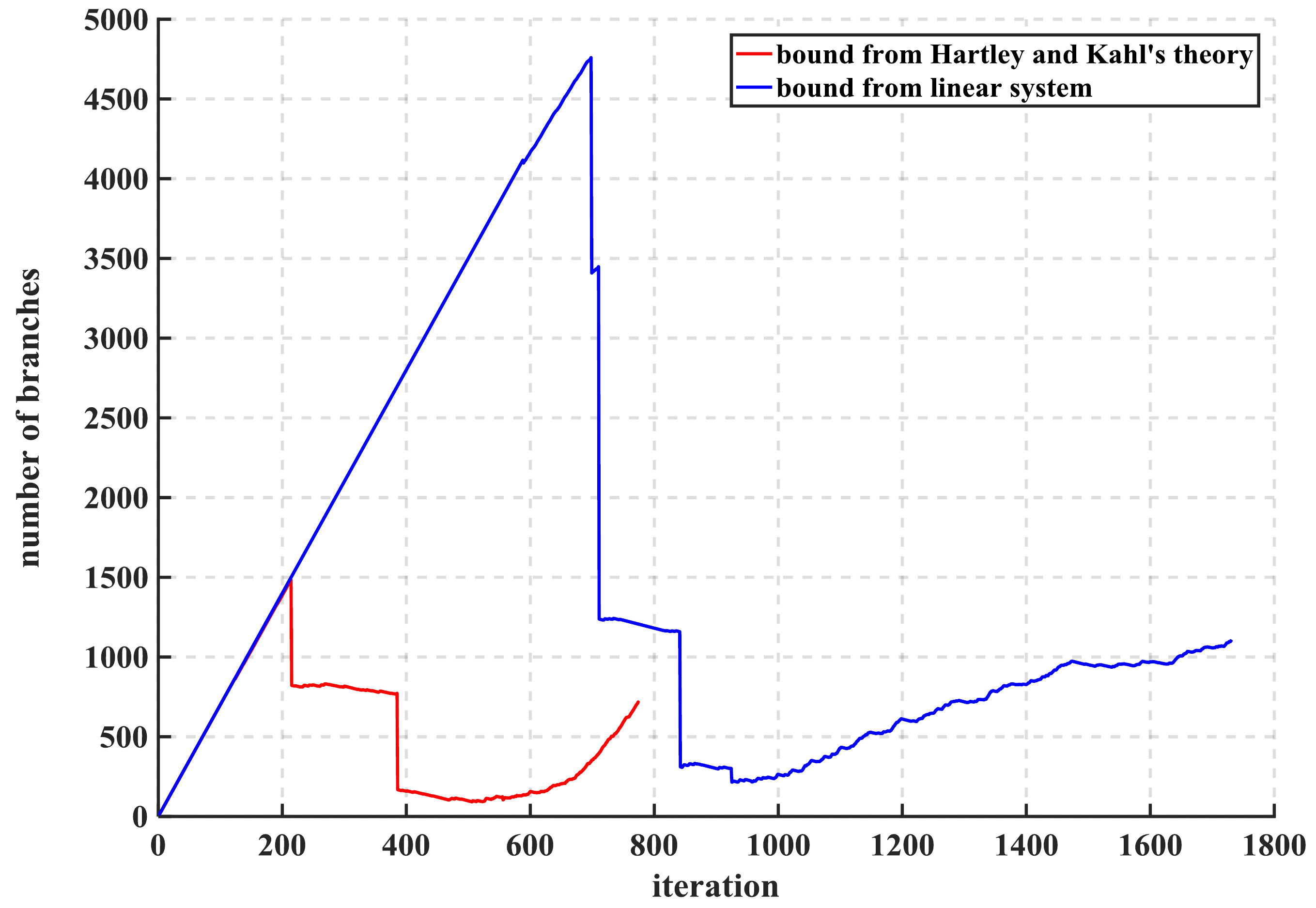}
\label{fig_second_case}}
\subfloat[]{\includegraphics[width=2.25in]{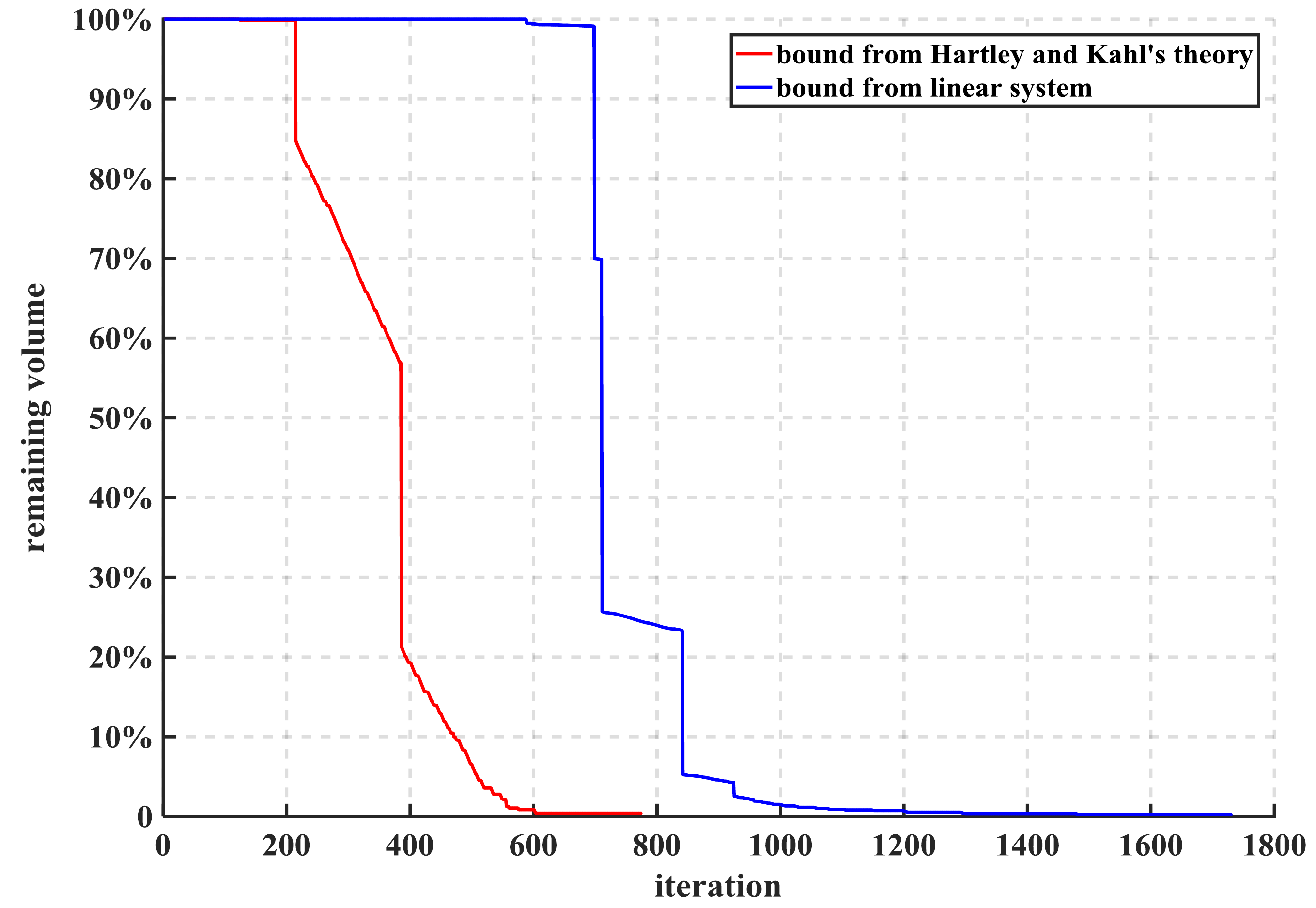}
\label{fig_third_case}}
\caption{ The optimality of RGPnP$\_$H and RGPnP$\_$L. From left to right: the evolution of the upper and lower bounds, the number of branches and the remaining volume.}
\end{figure*}

\begin{figure}[htbp]
\centering
\subfloat[]{\includegraphics[width=3.3in]{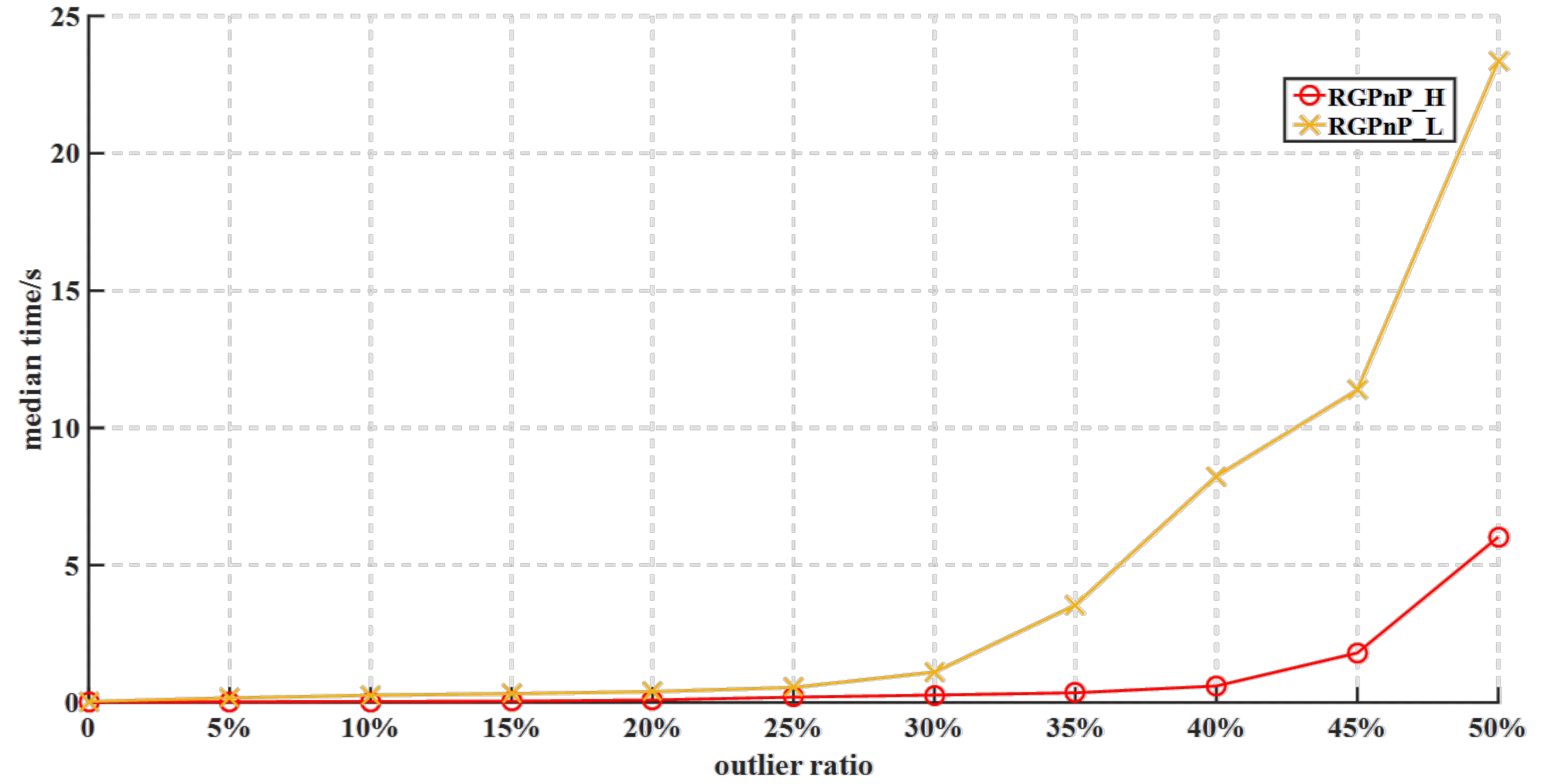}
\label{fig_first_case}}
\hfil
\subfloat[]{\includegraphics[width=3.3in]{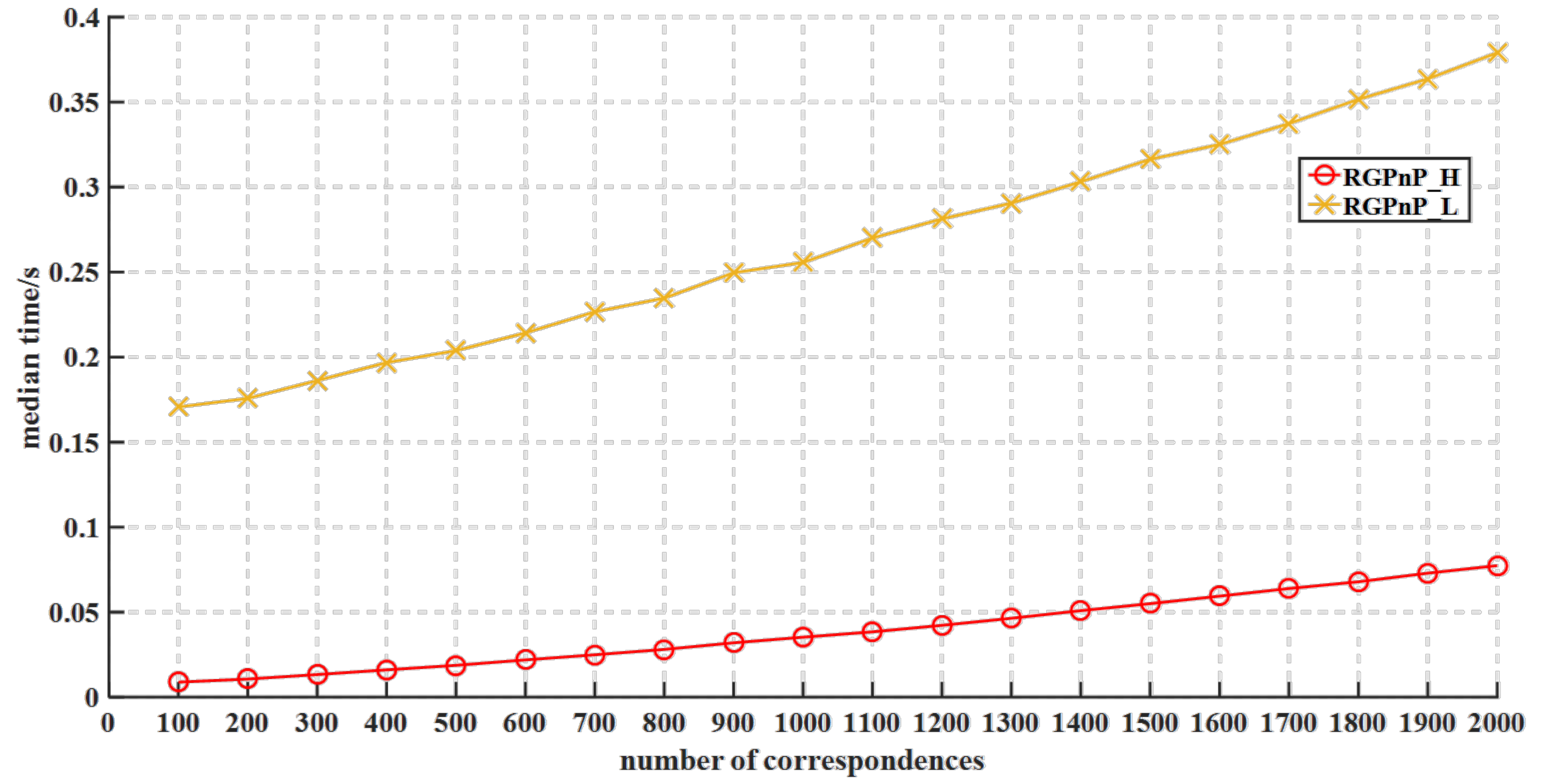}
\label{fig_second_case}}
\caption{The complexity and scalability of RGPnP$\_$H and RGPnP$\_$L. (a). Median run time versus the outlier ratio (with 1000 correspondences). (b). Median run time versus the number of correspondences (with 10$\%$ outliers).}
\end{figure}



As illustrated in Fig.3, our methods performed well in all trials with both types of outliers, while R1PPnP handled the first type of outliers well but failed on the second type. The RANSAC-based methods found the correct pose in most trials with a small outlier ratio but failed in most trials when the outlier ratio was large. REPPnP’s performance was unsatisfactory for both types of outliers. 



\textbf{Global optimality.} To demonstrate the global optimality of the proposed methods, we ran a trial with 25$\%$ outliers of the first type. We present the evolution of the upper and lower bounds, the number of branches and the remaining volume for each of our methods in Fig.4. The upper and lower bounds converged after 775 iterations and 1731 iterations for RGPnP$\_$H and RGPnP$\_$L, respectively, indicating that the bounds derived from Hartley and Kahl’s theory are tighter than those derived from the linear system formulation.

\textbf{Complexity and scalability.} In this section, we study the run time of the proposed method with respect to different outlier ratios and different numbers of correspondences. In the first experiment, there were 1000 correspondences in total, and we ran the two versions of the methods 500 times under different outlier ratios with outliers of the first type. The median run time among the 500 trials is shown in Fig.5(a). RGPnP$\_$H is faster than RGPnP$\_$L, and the median run time of RGPnP$\_$H is less than one second when the outlier ratio is no greater than 40$\%$. Then, we experimentally investigated the scalability of the two methods. We ran each of the two methods 500 times with different numbers of correspondences and 10$\%$ outliers of the first type, and the results are presented in Fig.5(b). Again, RGPnP$\_$H is faster than RGPnP$\_$L, and the run times of both methods increase linearly with respect to the number of correspondences. Even with 2000 correspondences, the median run time of RGPnP$\_$H is still less than 0.1 second. However, the run times are exponential in the outlier ratio, reflecting the hardness of the robust estimation problem.

\begin{figure*}[tbp]
\centering
\includegraphics[width=6.5in]{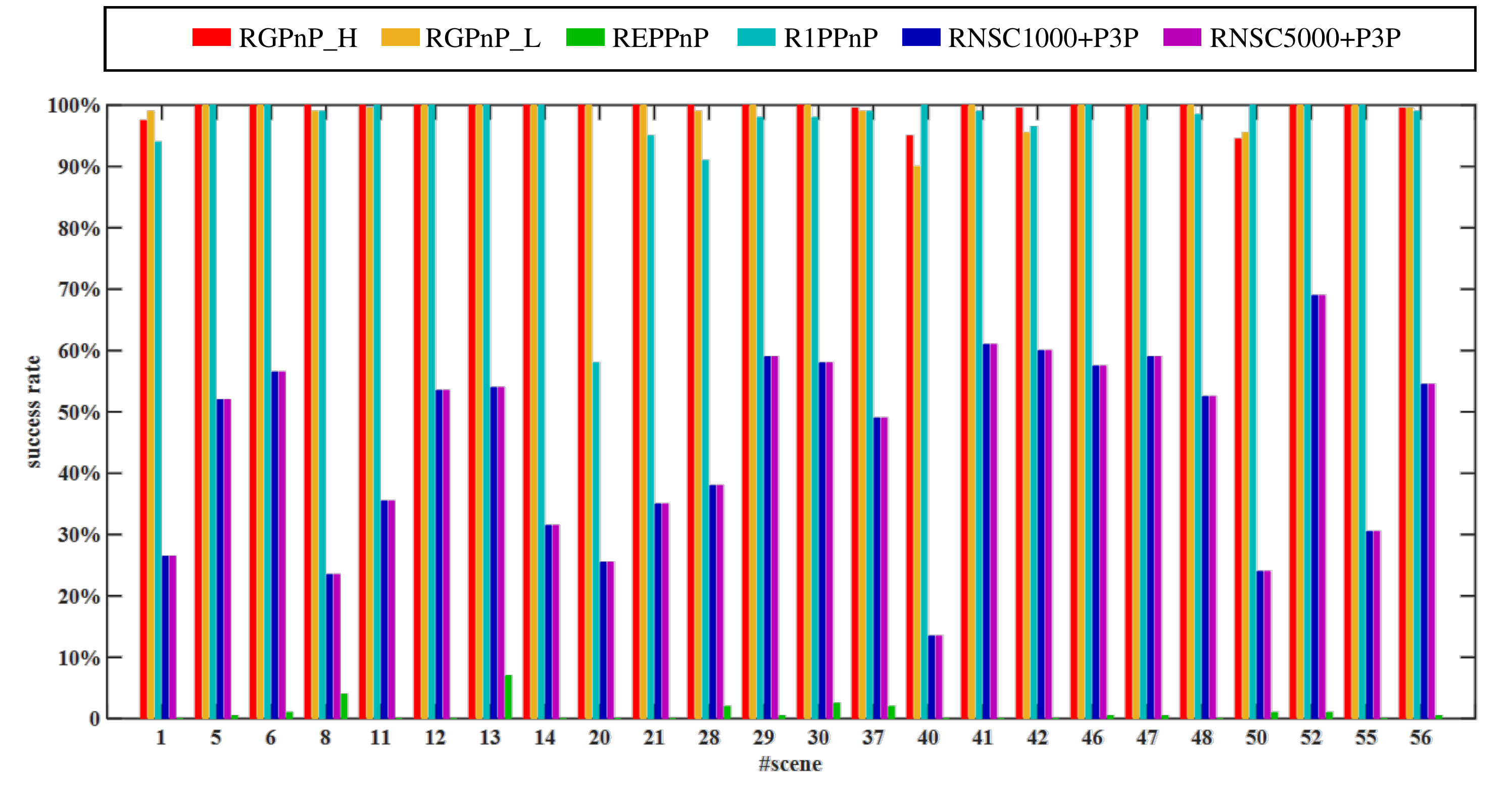}
   \caption{Success rates on the real-world data.}
\end{figure*}

\begin{figure*}[tbp]
\centering
\includegraphics[width=6.8in]{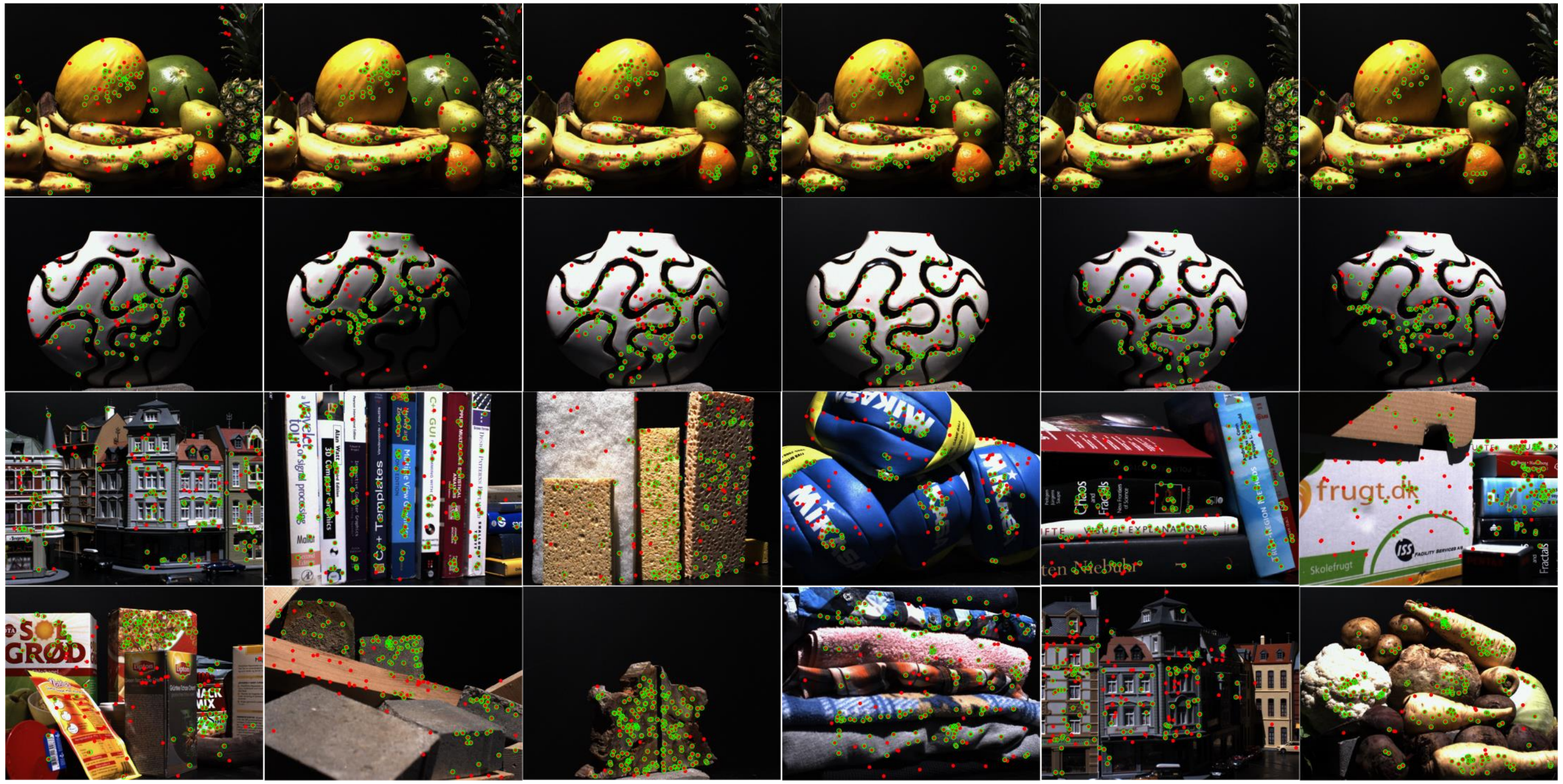}
   \caption{Examples of the real-world data. The red spots are all the correspondences detected based on SURF features, and the green circles are the inliers reprojected using RGPnP. First row: example images captured from different camera positions. Second row: example images captured under different illumination situations. Third and fourth rows: example images of different scenes.}
\end{figure*}

\subsection{Experiments with real-world data}
This section reports an evaluation conducted on the DTU Robot Image Data Sets \cite{Aanæs2012}. The data consist of images of 60 scenes of different kinds of objects and materials, each of which was captured from 119 camera positions under 19 illumination situations. The 3D point clouds were obtained by means of structured light scanning. The calibration information is provided and the resolution of the images is $1600\times1200$. For the experiment reported in this paper, 24 scenes were used. For each of these 24 scenes, we selected 20 camera positions and 10 illumination situations for each position, which resulted in a total of $24\times20\times10=4800$ 2D images. For each combination of scene and illumination situation, the image No.25 was used as the reference image, and we matched SURF features between the reference image and the images from each of the 20 camera positions considered in this experiment to create the correspondences between each feature point in the reference image and points in the other images. Then, we reprojected the related 3D point cloud onto the plane of the reference image to find the correspondences between the 3D points and the 2D SURF feature points in the reference image. In this way, we indirectly created 2D-3D correspondences for each of the 4800 2D images used in this experiment. The number of correspondences for each image ranged from 49 to 220. Note that the correspondences created in this way contained both outliers and noise and that the outlier ratio varied with scenes, camera positions and illumination situations. 

Then, we ran the six methods considered for comparison on all 4800 sets of 2D-3D correspondences and computed the estimation accuracy and the success rate as described in Sec 3.1. The results are presented in Fig.6 and Fig.7. Both versions of the proposed method achieved a 100$\%$ success rate for almost all scenes, camera positions and illumination situations. R1PPnP achieved results similar to those of our methods, while the other compared methods failed in most trials. These results indicate that our method produce the globally optimal solution and addresses outliers well. Fig.7 shows several examples of the real-world image data: after recovering the camera pose, we reprojected the 3D inliers onto the image plane with an inlier threshold of 10 pixels.

\section{Discussion and conclusion}
In this paper, we propose a novel method of solving the absolute pose estimation problem. Our method is robust to outliers in the 2D-3D correspondences, and it solves the problem in a globally optimal way, which means that our method is able to produce a guaranteed best solution. Specifically, we reduce the dimensionality of the original problem from six to three, which makes the branch-and-bound-based optimization process much faster. The 0.5n-subset can be used as the input when the outlier ratio is low; however, if the outlier ratio is high, which will greatly increase the run time, we recommend the common trick of applying a heuristic outlier removal method to significantly reduce the outlier ratio before using our globally optimal method. For our branch-and-bound algorithm for the rotation search, we propose two upper bounds: the first one is derived from Hartley and Kahl$'$s rotation search theory and is more efficient for our problem, whereas the other is an original contribution that is more general and could be extended for application to other problems.

{\small
\bibliographystyle{ieee}
\bibliography{mypaper_RGPnP_arxiv}
}

\end{document}